\documentclass{article}

\usepackage{arxiv}

\usepackage[utf8]{inputenc} 
\usepackage[T1]{fontenc}    
\usepackage{hyperref}       
\usepackage{url}            
\usepackage{booktabs}       
\usepackage{amsfonts}       
\usepackage{amsmath}
\usepackage{amssymb}
\usepackage{nicefrac}       
\usepackage{microtype}      
\usepackage{graphicx}
\usepackage{wrapfig}
\usepackage{floatflt}
\usepackage{algorithm}
\usepackage[noend]{algpseudocode}
\usepackage[innercaption]{sidecap}
\usepackage{subcaption}
\usepackage{threeparttable}

\title{ImmuNeCS: Neural Committee Search by an Artificial Immune System}

\author{
  Luc Frachon \thanks{Heriot-Watt University, UK} \\
  \texttt{ljf2000@hw.ac.uk} \\
   \And
  Wei Pang $^*$ \\
  \texttt{w.pang@hw.ac.uk} \\
   \And
  George M. Coghill \thanks{University of Aberdeen, UK}\\
  \texttt{g.coghill@abdn.ac.uk}
}

\begin{document}
\maketitle
\setcounter{footnote}{0}

\begin{abstract}
Current Neural Architecture Search (NAS) techniques can suffer from a few shortcomings, including high computational cost, excessive bias from the search space, conceptual complexity or uncertain empirical benefits over random search. In this paper, we present ImmuNeCS, an attempt at addressing these issues with a method that offers a simple, flexible, and efficient way of building deep learning models automatically, and we demonstrate its effectiveness in the context of convolutional neural networks. Instead of searching for the 1-best architecture for a given task, we focus on building a population of neural networks that are then ensembled into a neural network committee, an approach we dub \emph{Neural Committee Search}. To ensure sufficient performance from the committee, our search algorithm is based on an artificial immune system that balances individual performance with population diversity. This allows us to stop the search when accuracy starts to plateau, and to bridge the performance gap through ensembling. In order to justify our method, we first verify that the chosen search space exhibits the locality property. To further improve efficiency, we also combine partial evaluation, weight inheritance, and progressive search. First, experiments are run to verify the validity of these techniques. Then, preliminary experimental results on two popular computer vision benchmarks show that our method consistently outperforms random search and yields promising results within reasonable GPU budgets. An additional experiment also shows that ImmuNeCS's solutions transfer effectively to a more difficult task. We believe these findings can open the way for new, accessible alternatives to traditional NAS.
\end{abstract}


\section{Introduction}

Neural architecture search (NAS) has become one of the most active topics within deep learning. Its purpose is to develop algorithms that are able to automatically discover optimal neural network architectures for a given task. The main objective is to reduce the amount of time spent by human practitioners in designing architectures, thus saving costs or making deep learning more accessible to organizations and people lacking the expertise. Moreover, there are two additional potential benefits of using NAS: firstly, some specific use-cases might not be efficiently tackled with existing standard architectures from the literature and may therefore require an approach that is not biased by human priors. Secondly, many experiments show that NAS-based models often achieve extremely competitive performance on benchmarks while being more compact or efficient. This can find applications in resource-limited environments such as mobile devices.

While early works \cite{zoph2017, real2017} can outperform human-designed architectures on benchmark datasets, they required thousands of GPU-days and are therefore only accessible to a few organizations around the world. Therefore, one of the main challenges in NAS has become efficiency improvement, regardless of the search strategy employed. Many solutions have been proposed, examples of which include performance prediction \cite{domhan2015, zela2018, real2019, baker2017}, weight inheritance \cite{real2017, elsken2017, elsken2019b}, network morphism \cite{jin2019, elsken2019b, cai2018}, and parameter sharing \cite{saxena2016, pham2018, bender2018, liu2018, cai2019, xie2019}.

While they effectively improve efficiency, some of the efforts mentioned above also significantly increase the conceptual and algorithmic complexity of their associated methods. To some extent, this defeats the purpose of making deep learning more accessible, in so far as a user might want to understand the algorithm they are using and not simply execute a series of instructions.

Most recent NAS approaches \cite{real2019, liu2018, pham2018, elsken2019b} use a cell-based architecture search space, as per \cite{zoph2018, zhong2018}. While this approach dramatically reduces the size of the search space (see Section \ref{par:cellbased}), it is motivated by the human affinity for regular patterns, as seen in classical architectures such as ResNet \cite{he2016} and DenseNet \cite{huang2017}. To the best of our knowledge, there is no theoretical justification for such regular structures and a machine-led search might end up discovering better architectures if free from this prior.

The series of limitations mentioned above have led us to approach the NAS problem from a different angle, with the aim of providing a method that is simple, flexible, effective, and efficient. We first observe that typical NAS learning curves show rapid progress in the early phases of the search, followed by a long plateau period where only minor improvements are made. Most of the search time is spent in that plateau phase. Secondly, NAS methods typically search for a single architecture or cell, and discard the others. If instead, we retain the members of the final population and ensemble them into a neural network committee (NNC), we can gain an economical boost in the accuracy of our model. In turn, this allows us to stop the search earlier, towards the beginning of the plateau phase, and recover the missing performance by means of ensembling.

To enable such a gain, the members of the population must be individually competent, but also diverse in the classification errors they make \cite{brown2005, coelho2006, pasti2007, pasti2010, castro2011, zeng2014}. This is achieved by employing an artificial immune system (AIS), a class of bio-inspired algorithms that has proven to be able to find multiple high-quality local optima \cite{decastro2002a, decastro2002b}.

Another contribution of this work is that we verify three assumptions commonly made in the field of NAS research, but which are rarely verified: the locality property of the search space, the correlation between partial evaluation and final performance of architectures, and the validity of progressively growing neural networks, in other words the correlation between the performance of an architecture, and the performance of the same architecture with a slight increase in complexity.

The remainder of this paper is organized as follows: we first give a brief overview of the research in fields related to this work in Section \ref{sec:relatedwork}. In Section \ref{sec:descriptionofImmuNeCS}, we then describe ImmuNeCS (Immune-inspired Neural Committee Search) in detail, including the search space and search algorithm. Section \ref{sec:experimentsandresults} presents a series of experiments that we ran to 1) verify three common assumptions in NAS that are rarely validated; 2) assess the performance and efficiency of our method on two common datasets, and 3) evaluate the ability of an ImmuNeCS-produced population of neural nets to transfer to another task. We conclude by discussing the advantages and limitations of our approach and exploring future research directions.



\section{Related Work}
\label{sec:relatedwork}
In this section, we present some key advances in the field of NAS, particularly around efficiency improvement. We then give a brief overview of AIS algorithms and NNCs.

\subsection{Neural Architecture Search}
NAS is a field of deep learning that has attracted muchh attention over the last few years \cite{lindauer2019}. Early works \cite{zoph2017, real2017} achieved promising results on benchmark datasets at the cost of thousands of GPU-days. The promise shown by these papers triggered intensive research to try and improve the efficiency of NAS methods. Here, we describe some solutions that have come out of this general effort,  which can often be combined \footnote{We refer to \cite{elsken2019a} for a comprehensive review of NAS.}.

\paragraph{Cell-based search space}
\label{par:cellbased}
Given the complexity of deep neural networks, the search space can potentially be infinite, so all NAS approaches come up with some ways of reducing it to a more computationally tractable size. Inspired by state-of-the-art hand-crafted architectures \cite{he2016, szegedy2017, zagoruyko2016}, most recent NAS works search for motifs called \emph{cells} rather than full architectures, an idea first proposed in \cite{zoph2018, zhong2018}. These cells are repeated in a pre-defined way to generate models. This greatly reduces the search space and enables transferability to more complex tasks by simply increasing the number of assembled cells. It also allows performing the search using relatively small architectures, before "augmenting" the final model before training by increasing the number of cells. The speed-up associated with this form of architecture search is around one order of magnitude \cite{zoph2018}. 

However, cell-based search limits one's ability to explore one of the most intriguing aspects of NAS, namely its potential for discovering macro-architectures that humans would not have come up with. In this research, we choose not to follow this strategy and instead use a limited selection of low- or high-level blocks that the search algorithm can assemble in any way it deems most effective, similar to the work in \cite{suganuma2017}.

\paragraph{Weight sharing}
\label{par:parametersharing}
Some of the most significant gains in efficiency have been obtained by training only one large model and sampling subgraphs from it that share its weights \cite{saxena2016, pham2018, liu2018, cai2019, xie2019}. The approaches using this strategy routinely report speed-ups of two to three orders of magnitude. However, recent research \cite{sciuto2019, li2019} has cast some doubt over the superiority of weight sharing methods over a well-designed random search, because the performance ranking of subgraphs using the shared weights might not accurately represent the ranking of final, trained-from-scratch models.

\paragraph{Progressive search}
\label{par:progressivesearch}
Given a set $\mathcal{K}$ of possible operations for each of $L$ positions (i.e. layer or cell element), the number of possible architectures is $|\mathcal{K}|^L$. However, by making the search progressive, it is possible to achieve $\mathcal{O}(|\mathcal{K}|L)$: Starting from a trivial or small architecture, one can search among the $|\mathcal{K}|$ possible configurations for the next position, and repeat the process to incrementally grow the network. This approach is adopted in \cite{liuc2017, dong2018} in the context of a cell-based search and in \cite{byla2019} on full architectures. The progressive search method allows \cite{liuc2017} to report efficiency gains of 3 to 5 times in the number of models evaluated. 

Progressive search assumes that there is a better chance of obtaining a strong candidate when building up from an architecture that is already strong, an assumption that we test in this research (see Section \ref{ssec:assumptionvalidation}).

\paragraph{Partial evaluation}
\label{par:partialtraining}
In any NAS strategy, candidate evaluation is one of the key bottlenecks, as training each neural network can take several GPU-hours. As a result, a common solution to reduce this time is to partially evaluate candidates, i.e. train them  on a subset of the training set and/or for only a few epochs \cite{zoph2018, zela2018, real2019}. This is based on the assumption that there is strong rank-correlation between the accuracy of partially trained networks and that of the same architectures subjected to full training. This assumption is not always true in practice \cite{zela2018}; we therefore run an empirical verification in this paper before adopting this strategy.

\paragraph{Weight inheritance}
NAS methods generally derive new architectures from previously-evaluated ones, by applying some form of transformation. Weight inheritance is the idea that in such cases, the weights learned by the parent architectures are carried-over to the child candidate in all components (i.e. layers or cell elements) that have not been altered during the transformation process. The inherited weights can then be fine-tuned \cite{real2017}, or even frozen completely \cite{elsken2019b}. In this paper, we use weight inheritance as a starting point for fine-tuning.

Network morphism \cite{chen2016, wei2016, elsken2017, cai2018} is a more radical form of weight inheritance. Modifications to the architecture are designed to preserve the function that they represent, either exactly or approximately \cite{elsken2019b}, so that all the weights learned so far can be re-used.

Both forms of network morphism allow child architectures to be warm-started with weights that have been learned by their parent, thus reducing the evaluation time.

\subsection{Artificial Immune Systems}
\label{ssec:ais}
AIS algorithms are a class of evolutionary algorithms that date back from the early 2000's and are inspired by theories related to the mammal immune system \cite{burnet1978, jerne1974}. They rely on the idea of a population of antibodies that can proliferate and mutate depending on their affinity to detected antigens, while also interacting between themselves to maintain diversity. When an antibody has been able to bind to an antigen, it will enter a memory bank that speeds up and amplifies the immune response in case of a future encounter with the same antigen or one closely related to it (\emph{secondary response}).

In the context of optimization tasks, one of the simplest AIS algorithms is the Clonal Selection Algorithm or Clonalg \cite{decastro2002a}. It has shown its ability to locate multiple optima and implicitly account for multiple possible solutions. Subsequent developments have taken advantage of interactions between population members to promote diversity, leading to the Opt-AINet algorithm \cite{decastro2002b}. In this paper, we use a variant of Clonalg, with inspiration from Opt-AINet and provisions made for progressive search (see Section \ref{par:progressivesearch}). Compared to the Genetic Algorithm (GA) \cite{holland1992}, another popular population-based search method, AIS have the added benefit of not requiring crossover or recombination of candidates, i.e. the recombination of different parts of the ``genomes'' of two individuals to create an offspring. It is not obvious how one would define such a crossover operation: layers co-adapt during training, and simply combining a section of one candidate with a section of another candidate seems ill-suited to the neural network paradigm. The authors of \cite{real2017} seem to reach the same conclusion and forego crossover entirely, despite using an algorithm related to GA.

\subsection{Neural Network Committees}
Neural network committees are simply ensemble models using neural networks. As such, they have been studied for a long time \cite{hansen1990}, including in the context of convolutional neural networks (CNN) \cite{ciresan2011}, and can benefit from most of the research around ensembling strategies. The performance of an ensemble essentially depends on two factors: The quality of each model in the ensemble, and their error diversity. However, it is not always clear how to achieve this error diversity, or even how to define it \cite{brown2005}. A natural idea is to explicitly optimize models for accuracy and diversity \cite{zeng2014}. The approach in \cite{bochinski2017} promotes diversity by using a modified fitness function to discover a population of neural networks by an evolutionary algorithm. Their work presents similarities with ours, however the emphasis of their search algorithm is on general hyperparameters rather than complete architectures, and their experiments are limited to the MNIST dataset \cite{lecun1998} and simple models. 

The alternative to explicit diversity enforcement is to use a search algorithm that is inherently able to maintain diversity. It seems that AIS algorithms are effective in this regard \cite{pasti2007}. In particular, the authors of  \cite{pasti2010} compare Opt-AINet trained with accuracy as its sole objective, with another AIS (Omni-AINet) specifically designed to optimize for both accuracy and diversity. They show that the single-task Opt-AINet can outperform the multi-task setting. 

\section{Description of ImmuNeCS}
\label{sec:descriptionofImmuNeCS}
In this section, we first explain the observations that motivate our method, then detail the main components of ImmuNeCS.

\subsection{Key Observations}
\paragraph{Diminishing returns}
As with many optimization processes, NAS tends to exhibit a fast rate of improvement in the early phases of the search. However, progress gradually becomes more difficult and more and more time is required to achieve the required progress (regardless of the metric used to measure that goal). If we had at our disposal a method to economically boost the accuracy of our model(s), we might be able stop the search much earlier and still recover a similar level of performance.

\paragraph{Plurality of solutions} 
A common feature of all NAS methods is that they evaluate many candidates during the search, whose performance gradually improves. In the context of evolutionary algorithms for instance, this means that the final generation of candidates should all have relatively good performance. Most NAS methods retain the absolute best candidate based on some criterion (typically validation accuracy, or Pareto dominance in multi-objective search) and discard the knowledge accumulated by the others.

\paragraph{Neural Committee Search}
Motivated by the two observations above, we choose to follow a different approach to most existing NAS methods. Instead of discarding the final generation of network architectures, we ensemble them, which achieves a significant improvement in the prediction accuracy compared to even the best network in the population (typically close to 1\%pt in our experiments). This in turn allows us to use relatively aggressive termination criteria for the search (see Section \ref{ssec:searchbyais}), thus completing the whole process within a reasonable compute budget. This idea, which we dub \emph{Neural Committee Search} (NCS), shifts the NAS problem from focusing on a single architecture to growing a competent but diverse population of classifiers.

\subsection{Representation}
\label{ssec:representation}

\begin{SCfigure}[1.0]
\includegraphics[scale=.7]{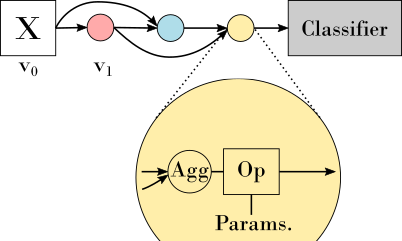}
\caption{Graph representation of an architecture with three hidden layers.}
\label{fig:representation}
\end{SCfigure}

We represent an architecture as a directed acyclic graph where nodes correspond to layers/blocks and edges to tensors (see Figure \ref{fig:representation}). Node $v_0$ is the start or input node, which receives data samples. Each subsequent node $v_l$ has at least one incoming edge, which comes from node $v_{l-1}$. On the CIFAR-10 \cite{krizhevsky2009} task, each hidden layer except $v_1$ can have a second incoming edge from any node $v_{l-k}, \, k>1$, thus allowing skip connections. The start node is the only node that is allowed an outdegree greater than 2. On the Fashion-MNIST \cite{xiao2017} task, all nodes have indegree 1.

Each node has an \emph{Aggregation} and an \emph{Operation}. Aggregations are used to combine tensors from earlier layers. They are \emph{None} when the node has an indegree of 1 but can be \emph{Add} or \emph{Concatenate} when it is 2, as described in Section \ref{par:cifar10searchspace}. Operations apply some mathematical transformation to their input tensor (e.g. convolution, ResNet block...). As per Table \ref{tab:searchspace}, each Operation can have several hyperparameters.

\subsection{Search Space}
The search space of any NAS method requires careful consideration. Too permissive, the dimensionality becomes unmanageable and the search becomes untractable. Too restrictive, and the algorithm may be prevented from discovering the interesting architectures. Any search space design will necessarily include some human bias in the choice of operations that are accessible, the hyperparameters of these operations, or the ways they can be connected to one another. However, as stated in Section \ref{par:cellbased}, we would still like the search to be able to surprise us with solutions that humans would not have thought of. In this research, we design two different search spaces depending on the dataset used in the experiments. We chose to design relatively small search spaces, but with solutions that might appear unconventional compared to other NAS methods.

\paragraph{Fashion-MNIST search space}
\label{par:fashionmnistsearchspace}
Fashion-MNIST is a dataset of small, $28 \times 28$-pixel greyscale pictures of ten classes of clothing items. It is a good choice for developing and testing a classification model because it is simple enough that our method can complete the search in about one day on two NVidia RTX 2080Ti GPUs, yet complex enough that there can be significant differences in final accuracy between different experimental settings. On this task, we restrict the search space to strictly sequential architectures (no skip connections). Each layer can be a convolution, a depthwise separable convolution, a pooling layer or an Identity function \footnote{The raison d'\^etre of the identity function is to possibly neutralize an existing layer during subsequent mutations, see Section \ref{par:cloneandmutate}.}. Each of these operations except Identity has three hyperparameters with a number of discrete values that they can take (see Table \ref{tab:searchspace}). Note that we deliberately make two decisions:

\begin{itemize}
    \item Somewhat contrary to common practice, \emph{Conv} and \emph{DSepConv} layers do not necessarily include batch normalization nor an activation function. In practice, the algorithm does indeed make use of this freedom. 
    \item For simplicity, pooling layers are solely responsible for widening via their \emph{Channel multiplier} parameter. Other layers do not change the number of channels.
\end{itemize}

The input layer takes data batches and applies a pointwise convolution to them to increase the channel count from 1 to 64, with no batch normalization nor activation. The network's final layer, the classifier, is also fixed and follows a common structure in CNNs. It first applies adaptive global concatenation pooling, whose number of channels is twice the output of the feature extraction part of network \cite{howard2019}. Then the data goes through batch normalization and dropout. The dropout rate is set at 20\% based on side experiments. Finally, a fully-connected layer reduces the number of activations to the number of classes of the task at hand (10).

\paragraph{CIFAR-10 search space}
\label{par:cifar10searchspace}
CIFAR-10 is one of the most popular computer vision tasks in recent NAS research. It consists of small, $32 \times 32$-pixel color images, distributed across ten classes. It is a much more challenging task than Fashion-MNIST and requires deeper architectures with skip connections. Letting the algorithm incrementally assemble simple layers as with Fashion-MNIST would be very slow, yet for reasons explained in Section \ref{par:cellbased}, we want to avoid the cell-based search paradigm.

Inspired by \cite{suganuma2017}, we therefore decide to design the search space around a menu of high-level blocks. These are taken from the classical deep learning literature \cite{he2016, huang2017, szegedy2017} but we do not impose any restrictions on the order nor the number of each block in the final architecture. As before, each block type has hyperparameters that the AIS can choose from (see Table \ref{tab:searchspace}). Here again, the input layer is simply a pointwise convolution with an output width of 32. The classifier part is the same as with Fashion-MNIST.

\begin{table}[t]
\resizebox{\textwidth}{!}{%
    \centering
    \begin{tabular}{lccc}
    \toprule
    \textbf{Operation Type} & 
        \begin{tabular}[c]{@{}c@{}}\textbf{Hyperparameter 1}\\ \textbf{\{Possible values\}}\end{tabular} &
        \begin{tabular}[c]{@{}c@{}}\textbf{Hyperparameter 2}\\ \textbf{\{Possible values\}}\end{tabular} &
        \begin{tabular}[c]{@{}c@{}}\textbf{Hyperparameter 3}\\ \textbf{\{Possible values\}}\end{tabular} \\
    \midrule
    Conv & \begin{tabular}[c]{@{}c@{}}Kernel size\\ \{1,3,5,7\}\end{tabular} & \begin{tabular}[c]{@{}c@{}}BatchNorm\\ \{yes/no\}\end{tabular} & \begin{tabular}[c]{@{}c@{}}ReLU\\ \{yes/no\}\end{tabular} \\ \midrule
    DSepConv & \begin{tabular}[c]{@{}c@{}}Kernel size\\ \{1,3,5,7\}\end{tabular} & \begin{tabular}[c]{@{}c@{}}BatchNorm\\ \{yes/no\}\end{tabular} & \begin{tabular}[c]{@{}c@{}}ReLU\\ \{yes/no\}\end{tabular} \\ \midrule
    Pool & \begin{tabular}[c]{@{}c@{}}Type\\ \{Max, Avg\}\end{tabular} & \begin{tabular}[c]{@{}c@{}}Kernel size\\ \{3,5\}\end{tabular} & \begin{tabular}[c]{@{}c@{}}Ch.multiplier\\ \{1, $1\frac{1}{3}$, $1\frac{2}{3}$, 2\}\end{tabular} \\ \midrule
    Identity & -- & -- & -- \\ 
    \bottomrule
    \end{tabular}
    \quad
    \begin{tabular}{lcc}
    \toprule
    \textbf{Operation Type} & 
        \begin{tabular}[c]{@{}c@{}}\textbf{Hyperparameter 1}\\ \textbf{\{Possible values\}}\end{tabular} &
        \begin{tabular}[c]{@{}c@{}}\textbf{Hyperparameter 2}\\ \textbf{\{Possible values\}}\end{tabular} \\
    \midrule
    Resnet Block\cite{he2016} & 
        \begin{tabular}[c]{@{}c@{}}Kernel size\\ \{3,5\}\end{tabular} & 
        \begin{tabular}[c]{@{}c@{}}Downsample\\ \{yes/no\}\end{tabular} \\
    \midrule
    \begin{tabular}[l]{@{}l@{}}Resnet Bottleneck\\ Block\cite{he2016}\end{tabular} & 
        \begin{tabular}[c]{@{}c@{}}Kernel size\\ \{3,5\}\end{tabular} &
        \begin{tabular}[c]{@{}c@{}}Downsample\\ \{yes/no\}\end{tabular} \\ 
    \midrule
    Densenet Block\cite{huang2017} & 
        \begin{tabular}[c]{@{}c@{}}Growth factor\\ \{12, 24, 36\}\end{tabular} & 
        \begin{tabular}[c]{@{}c@{}}Transition layer\\ \{yes/no\}\end{tabular} \\
    \midrule
    \begin{tabular}[l]{@{}l@{}}Densenet Bottleneck\\ Block\cite{huang2017}\end{tabular}  & 
        \begin{tabular}[c]{@{}c@{}}Growth factor\\ \{12, 24, 36\}\end{tabular} & 
        \begin{tabular}[c]{@{}c@{}}Transition layer\\ \{yes/no\}\end{tabular} \\
    \midrule
    \begin{tabular}[l]{@{}l@{}}Inception-Resnet\\ Block A\cite{szegedy2017}\end{tabular}  & 
        \begin{tabular}[c]{@{}c@{}}Kernel size\\ \{3, 5\}\end{tabular} & 
        \begin{tabular}[c]{@{}c@{}}Bottleneck factor\\ \{0.1, 0.4, 0.75\}\end{tabular} \\
    \midrule
    \begin{tabular}[l]{@{}l@{}}Inception-Resnet\\ Block B\cite{szegedy2017}\end{tabular}  & 
        \begin{tabular}[c]{@{}c@{}}Kernel size\\ \{3, 5\}\end{tabular} & 
        \begin{tabular}[c]{@{}c@{}}Bottleneck factor\\ \{0.1, 0.4, 0.75\}\end{tabular} \\
    \midrule        
    Pool & 
        \begin{tabular}[c]{@{}c@{}}Type\\ \{Max, Avg\}\end{tabular} &       
        \begin{tabular}[c]{@{}c@{}}Kernel size\\ \{3,5\}\end{tabular} \\ 
    \midrule        
    Identity 
        & --
        & -- \\ 
    \bottomrule
    \end{tabular}
}
\caption{Operations, hyperparameters and their possible values included in the Fashion-MNIST (left) and CIFAR-10 (right) search spaces. \emph{Ch.multiplier} refers to the factor by which pooling layers increase the number of channels. \emph{Downsample} indicates whether the block should reduce the spatial size of the activations by a factor 2 and increase the channel count by the same factor. \emph{Growth factor} is the number of channels that are added to the data tensors as they go through the block. \emph{Transition layer} indicates whether a compression layer is appended to reduce the channel count after the block by 50\%. \emph{Bottleneck factor} is an internal compression factor in the channel count that affects the middle branch of the block. The output of the block retains the same number of channels as its input. See respective papers for details of the blocks.}
\label{tab:searchspace}
\end{table}

Each node $v_l$ is decoded into a block function $B_l$ which receives a tensor $\xi_{l-1}$ from block $B_{l-1}$, and optionally, a tensor $\xi_k$ from any block $B_k, \, 0 \leq k < l-1$. $\xi_{l-1}$ and $\xi_k$ can be either summed up, or concatenated along the channel dimension. Mismatches between spatial dimensions are resolved by bilinear interpolation, whereas channel counts are aligned by pointwise convolutions without batch normalization nor activation function. The output from $B_l$ can thus be:
\begin{itemize}
    \item $B_l(\xi_{l-1})$ if $B_l$ has only one input (which is only certain for $B_1$),
    \item $B_l(\xi_{k} + \xi_{l-1})$ or $B_l(\xi_{k} \oplus \xi_{l-1}), \, 0 \leq k < l-1$ otherwise (where $\oplus$ denotes concatenation along the channel dimension, ignoring spatial and width adjustments with slight abuse of notation).
\end{itemize}

Unlike most recent works in NAS, we do not predefine where the data should be downsampled or the number of channels increased in the macro-architecture. Instead, these manipulations are freely decided by the search algorithm. In theory, and given enough network evaluations, this allows the algorithm to adapt the receptive field and number of channels to local requirements within each architecture, rather than mandating a cell structure that can work around fixed values for both these structural decisions.

\subsection{Search by an AIS} 
\label{ssec:searchbyais}
We first give an overview of our Clonalg-derived search algorithm, before detailing its key components: cloning and mutation, random insertions, and augmentation.

\paragraph{Overview}
The search is conducted by an AIS, starting from a population of $N$ networks comprising of a small number of random hidden layers. Then the networks go through partial training and are evaluated on a validation set. Their validation accuracy represents their \emph{affinity} to the task. 

Subsequently, at every generation, a fixed number of clones is generated for each network. These clones undergo mutations to their connections, aggregation and operation (see details below). The resulting architectures are trained and evaluated, and the best $N$ candidates from the pool of parents and clones are retained to form the next generation. The mean affinity of the population is then computed, and a few new random networks are inserted into the population to explore new regions of the search space. If the mean affinity does not improve by more than a threshold $\tau$ within a patience period of $\pi$ generations, the population goes through \emph{augmention}: The algorithm generates clones of each individual and appends one random layer to each of them. This progressive search mechanism allows the AIS to generate minimal networks for each task. A layer can be modified through mutation even after the addition of subsequent layers, which prevents models from being locked in the sub-region of the search space defined by the previous layers.

The whole process then restarts from the cloning and mutation step. The search terminates when $\pi$ consecutive augmentation phases have not yielded sufficient improvement, i.e. the mean affinity has not improved by more than $\tau$. Note that these two hyperparameters control the threshold and patience for both the inner loop (how many generations to wait for an improvement before augmenting the population), and the outer loop (how many augmentations to wait for an improvement before stopping the search). By changing the values of $\pi$ and $\tau$, one can make the search stop earlier or later in the learning curve. In our experiments, we used $\pi=2$. $\tau$ was usually set at 0.0075 for Fashion-MNIST and 0.003 for CIFAR-10, which correspond to relatively steep parts of the respective learning curves (0.75/0.3\%pt improvement in mean accuracy within two generations).  The pseudo-code is provided in Algorithm \ref{alg:ais}.

\begin{algorithm}[t]
\caption{ImmuNeCS Search} 
\begin{algorithmic}
\Procedure{SEARCH}{$N$: population size, $L_{ini}$: initial number of layers, $\rho$: mutation factor, $n_c$: number of clones per parent, $N_i$: number of random insertions, $n_a$: number of augmented networks per parent, $\pi$: patience, $\tau$: threshold}
\State pop $\gets$ MakeRandomArchitectures($N$)
\State pop $\twoheadleftarrow$ MakeAugmentedCopies(pop, $n_a$)  \Comment{Population size = $N(1 + n_a$)}
\State Evaluate(pop)  \Comment{Assign affinity $\operatorname{f}(i)$ to each network $i$}
\While{ExitCondition($\pi$, $\tau$) not met}: 
    \State Make training and validation datasets
    \State clones $\gets$ Clone(pop, $n_c$) \Comment{$N(1+n_a)n_c$ clones in total}
    \State clones $\gets$ MUTATE(clones, $\rho$)  \Comment{See Algorithm \ref{alg:mutate}}
    \State Evaluate(clones)
    \State pop $\gets$ SelectNBest(pop $\cup$ clones, $N$)  \Comment{Population size = N}
    \State avg\_affinity $\gets$ ComputeAverageAffinity(pop)
    \State pop $\twoheadleftarrow$ MakeRandomArchitectures($N_i$)  \Comment{Population size = $N + N_i$, $N_i$ small (1 or 2)}
    \State Evaluate($\text{pop}_{\{N, \dots, N + N_i\}}$)  \Comment{Train and evaluate new architectures only}
    \If{\emph{avg\_affinity} improved by less than $\tau$ for $\pi$ generations}:
            \State pop $\twoheadleftarrow$ MakeAugmentedCopies(pop, $n_a$)  \Comment{Population size = $(N + N_i)(1 + n_a)$}
            \State Evaluate($\text{pop}_{\{N + N_i, \dots, (N + N_i)(1 + n_a)\}})$  \Comment{Train and evaluate new architectures only}
    \EndIf
\EndWhile
\Return pop
\EndProcedure
\end{algorithmic}
\label{alg:ais}
\end{algorithm}

\paragraph{Cloning and mutation}
\label{par:cloneandmutate}
The mutation strategy is an important difference of AISs compared to traditional evolutionary methods such as the genetic algorithm. Here, the magnitude of the mutations that clones undergo is influenced by the affinity of their respective parents. This is a way of balancing exploitation and exploration: if a parent has high affinity, the AIS will focus on a small region around this parent. Conversely, when a parent has poor affinity, the AIS will extend the search to a wider area in an attempt to find more promising solutions. In addition, as shown in Equation (\ref{eq:mutationstrength}), we assign linearly larger mutation variances to more recent layers, i.e. closer to the network's head. This is because earlier layers have already had several opportunities to mutate and we do not want to generate more mutations than required, as they are costly in terms of re-training the network.

\begin{wrapfigure}{r}{0.4\textwidth}
\vspace{-0.5cm}
\centering
\includegraphics[width=0.9\linewidth]{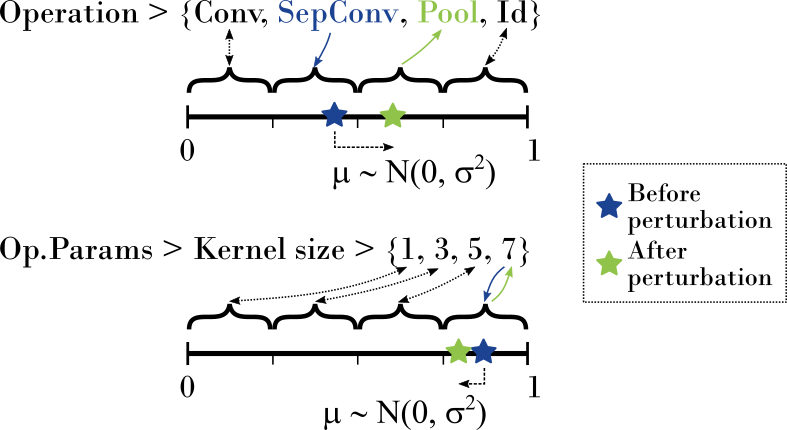}
\caption{Perturbation of continuous hyperparameter values and binning into discrete values.  \textbf{Top example}: the strength of the perturbation was sufficient to change the discretized value of the operation type.  \textbf{Bottom example}: the perturbation was too small to change the expression of the kernel size hyperparameter, but the change in its continuous value is recorded and becomes the starting point for future mutations.}
\label{fig:perturb}
\end{wrapfigure}

In practice, given a clone of depth $L$ (i.e. with $L$ layers/building blocks), the mutation rate $\alpha$ is computed as:
\begin{equation}
  \alpha=e^{-\frac{\operatorname{f_{parent}}}{\rho}}   
\end{equation}
with $\operatorname{f_{parent}}$ being the affinity of the clone's parent architecture. Then for the $l^{\text{th}}$ layer or block, the strength of the mutation $\mu$ is randomly sampled as:

\begin{equation}
  \mu \sim \mathcal{N}(0, \sigma^2) \;  \text{where} \; \sigma=\alpha \frac{l+1}{L}
  \label{eq:mutationstrength}
\end{equation} 

However, the mutation operator above assumes continuous values, whereas we are dealing with discrete features conditioned by other features. To resolve this issue, we store continuous values in the range $[0,1]$ for each hyperparameter and discretize them on the fly through binning when constructing each network (see Figure \ref{fig:perturb}). This allows us to remember all mutations even when they were insufficient to alter the integer value of a hyperparameter (a similar strategy was employed in \cite{real2017}).

\begin{algorithm}[t]
\caption{Mutation operator pseudo-code} 
\begin{algorithmic}
\Procedure{MUTATE}{clones, $\alpha$: mutation factor }
\For{each clone $c_j$ with parent's fitness $\mathrm{f}_j$}:
    \For{each node (layer) $v_l$, $l>1$,}:  \Comment{Mutate connections, CIFAR-10 only}
        \State PERTURB $v_l$'s indegree  \Comment{see explanation of PERTURB on Figure \ref{fig:perturb}}
        \If{$v_l$'s indegree == 2}: 
            \State \textbf{if} indegree was 1 before, sample a node as $2^{\text{nd}}$ input   \Comment{Uniform random from predecessor nodes}
            \State \textbf{else} PERTURB $2^{\text{nd}}$ input node's index  \Comment{The $1^{\text{st}}$ input is always from $v_{l-1}$}
        \EndIf
    \EndFor
    \If{$c_j$'s connections are unchanged}:  \Comment{Assessed after discretization}
        \For {each node (layer) $v_l$, $l>0$,}:  \Comment{Mutate aggregations and operations}
            \State \textbf{if} $v_l$'s indegree == 2, PERTURB Aggregation type  \Comment{Add or Concat}
            \If{(discrete) Aggregation type has not changed}: 
                \State PERTURB Operation type
                \If{(discrete) Operation type has changed}:
                    \State Sample new Operation hyperparameters  \Comment{Uniform random from [0,1]}
                \Else
                    \State PERTURB each Operation hyperparameter
                \EndIf
            \EndIf
        \EndFor
    \EndIf
\EndFor
\Return clones
\EndProcedure
\end{algorithmic}
\label{alg:mutate}
\end{algorithm}

Each clone's mutation sequence is described in Algorithm \ref{alg:mutate}. To avoid changes too drastic, we do not let mutations affect both the connectivity of the clone and its layers, so if at least one connection has changed, we immediately move on to the next clone. Otherwise, we start mutating nodes: for each of them, we first perturb the aggregation type (if applicable, i.e. the node has two incoming edges). Next, we try to perturb the node's operation type, then all of the node's operation hyperparameters. All these perturbations use Equation (\ref{eq:mutationstrength}), and at each step, we assess whether the discretized value has changed as per Figure \ref{fig:perturb}. If it has changed, we skip the next steps and move on to the next clone. By proceeding in this way, we ensure that only small elements of the networks are changed at every round of mutation, ensuring consistency between the parents' and clones' performance (see the experiment on locality and mutation operation in Section \ref{ssec:assumptionvalidation}).

To fully utilize the population's capacity, we keep track of all architectures already evaluated using a unique string encoding scheme and do not allow mutated clones to be identical to previously encountered networks. All mutated clones inherit the weights learned by their parent, except in the layers that have been modified by mutations (as assessed \emph{after} discretization), or whose number of channels has changed due to mutations to upstream components. Layers that do not inherit weights are initialized with He initialization \cite{he2015}.

\paragraph{Random insertions}
Novelty introduction is a way to promote exploration during the search. In our method, it is also a way of improving diversity, which as we saw is essential to NNCs. Other evolutionary algorithms frequently resort to crossover operations. However, as argued is Section \ref{ssec:ais}, such approaches seem ill-suited when it comes to neural networks. Instead, we simply insert a small number of random individuals into the population at every generation. To make sure that their capacity is comparable to the rest of the population and therefore, to give them a chance to survive the next selection operation, their depth is set at the current average depth of the population.

\begin{figure}[t]
    \centering
    \includegraphics[width=\textwidth]{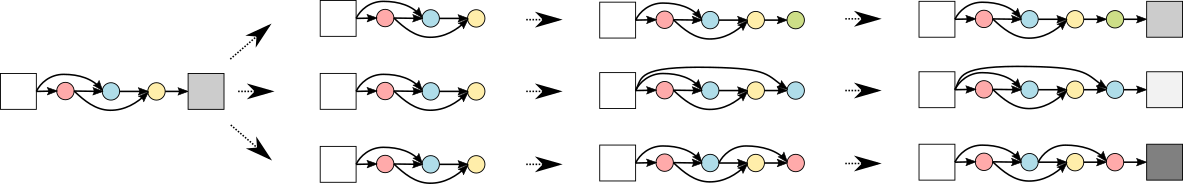}
    \caption{Network augmentation strategy for progressive search.}
    \label{fig:augmentation}
\end{figure}

\paragraph{Network augmentations}
As mentioned in Section \ref{par:progressivesearch}, we adopt a progressive search strategy. To this end, we periodically increase the capacity of the population by cloning all networks, removing the heads of all clones, appending one random layer to each (with the number of incoming edges sampled randomly), and rebuilding their heads (see Figure \ref{fig:augmentation}). It is worth noting that the original, non-augmented networks remain part of the population so that augmentations that do not bring any benefits can be ignored. As with mutations, clones inherit the weights learned by their parent, except for the new layer, which is initialized with He initialization.

\paragraph{Evaluation during search}
\label{par:evaluation}
Network graphs are only decoded into the actual neural network architectures at evaluation time or when inheriting weights from their parent. Otherwise, we only keep the graph representations in memory. As the training and evaluation of candidate solutions is by far the most resource-hungry part of the algorithm, we resort to partial evaluation. Each candidate is trained on only 20\% of the training set, and we define an aggressive early stopping policy: If the validation accuracy does not improve by more than 0.5\% on Fashion-MNIST (0.3\% on CIFAR-10) for 2 epochs, training stops. In all cases, training is not allowed to exceed 15 epochs on Fashion-MNIST and 30 on CIFAR-10. At that point, the weights of the best epoch seen so far are saved, and the network is assigned the best validation accuracy as its affinity score. In addition to saving time, this strategy favors architectures that train fast.

Driven by the objective of improving the whole population, we monitor progress by computing the average population affinity at every generation, rather than the single best affinity as is typically the case in NAS. Note that we attempted more complex methods to explicitly promote diversity, such as the modified fitness function used in \cite{bochinski2017}, or deleting architectures whose prediction errors were too similar to others (inspired by Opt-AINet \cite{decastro2002b}), but have found no obvious benefits from either method. In other words, the AIS's intrinsic ability to maintain diversity was found to be enough. We therefore decide to retain the simpler metric.

\subsection{Neural committee building}
Once the search is completed, we are left with $N$ partially trained architectures. On the Fashion-MNIST task, all architectures are retained. On CIFAR-10, due to the longer time required to train each model, we retain the best $\frac{N}{3}$ candidates based on their affinity scores. We train the retained networks further without reinitializing their weights, on the full training set and for more epochs (see experimental details in Section  \ref{sec:experimentsandresults}).

The NNC's class predictions are obtained by \emph{weighted soft majority vote}. Given a data sample $\mathbf{X}_i$ and a committee made up of $N_{\mathrm{NNC}}$ neural nets $g_j$ with affinity scores $\operatorname{f}_j$ ($j \in \{1, \dots, N_{\mathrm{NNC}}\}$), we get a collection of $c$-dimensional probability vectors $g_j(\mathbf{X}_i)$, where $c$ is the number of classes in our classification task. We then compute the following weighted sum:
$$
G(\mathbf{X}_i) \doteq \sum_{j=1}^{N_{\mathrm{NNC}}}\left( \frac{\operatorname{f}_j}{\sum_{m=0}^{N_{\mathrm{NNC}}}\operatorname{f}_m} g_j(\mathbf{X}_i) \right)
$$

Finally, the committee's prediction for this sample is given by $\operatorname{argmax} G(\mathbf{X}_i)$.

\section{Experiments and Results}
\label{sec:experimentsandresults}
In this section, we first describe experiments that were conducted to verify a number of common assumptions in NAS. We then present preliminary results from full-scale experiments on the two tasks and compare them to several existing NAS methods. Finally, we run a comparison to random search to assert that the performance of our method does not only result from the search space design.

\subsection{Assumption Validation}
\label{ssec:assumptionvalidation}
A central, although implicit, assumption made by NAS is that the search space of all architectures exhibits the locality property: Architectures that are nearby in the hyperparameter space will perform similarly. However, this is not an assumption that is commonly tested, so we run an experiment to verify it, and at the same time verify that our mutation operator is defined in a way that exploits this locality property. Moreover, as described in Section \ref{ssec:searchbyais}, we employ a few tricks to speed up the search: weight inheritance, partial evaluation, and progressive network growth. Weight inheritance is a common practice in many papers \cite{real2017, elsken2019b} and its validity is verified in \cite{elsken2017}. However, the other two techniques require further validation. 

\paragraph{Locality and mutation operator}
\label{par:assval_locality}
To test for locality, we generate a population of 100 random networks in the Fashion-MNIST search space. To test the assumption for both shallow and deep networks, half the population has depth 3 and the other half, depth 9. Each network is evaluated against the Fashion-MNIST task, then 10 mutated clones are generated for each parent, using the mutation operator described in Section \ref{par:cloneandmutate}. For each parent, we compute the mean and standard deviation of its clones' affinity scores and analyse their correlation to the parent's affinity.

If the locality assumption holds, and if the mutation operator can take advantage of it, we can expect two things to happen. Firstly, the mean affinity of the clones should be correlated to the parent's affinity. Secondly, as the mutation operator applies mutations with a larger variance to clones whose parent has a low affinity score, we should observe an inverse correlation between parent affinity and the variance in clone affinity.

\begin{figure}
\centering
\begin{subfigure}{0.48\textwidth}
    \centering
    \includegraphics[width=\linewidth]{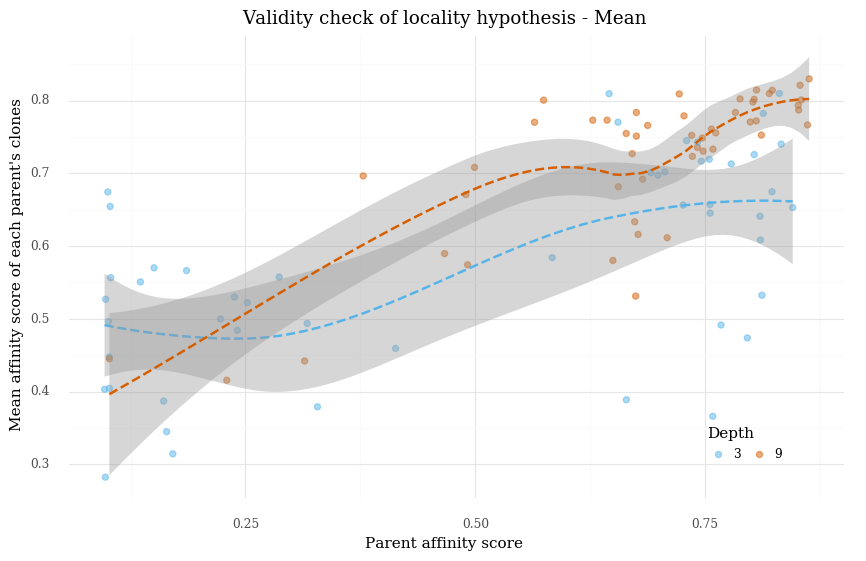}
    \subcaption{Spearman rank correlation for depth = 3:\\ $r = 0.53, p < 10^{-4}$; depth = 9: $r = 0.70, p < 10^{-9}$}
\end{subfigure}%
\begin{subfigure}{0.48\textwidth}
    \centering
    \includegraphics[width=\linewidth]{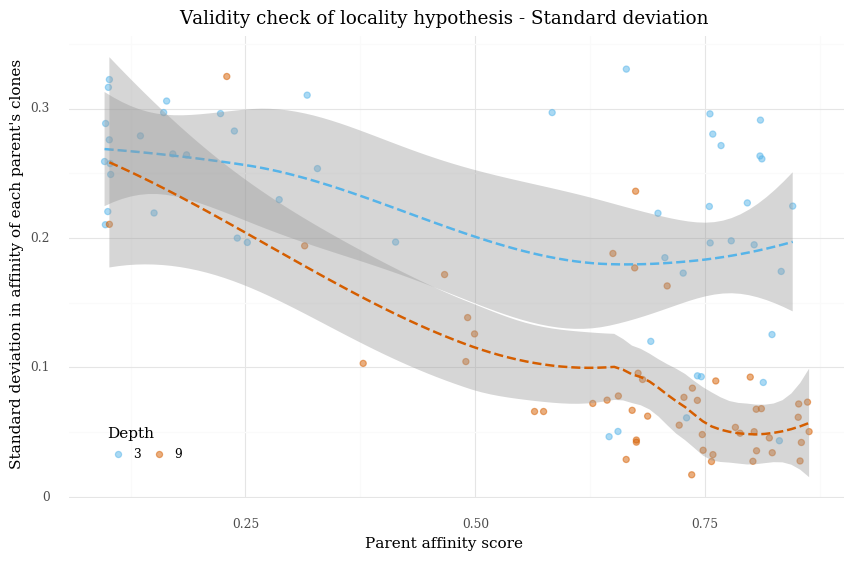}
    \subcaption{Spearman rank correlation for depth = 3:\\ $r = -0.41, p < 0.01$; depth = 9: $r = -0.60, p < 10^{-5}$}
\end{subfigure}
\caption{Validity check of the locality assumption, using mean (a) and standard deviation (b) of clones vs their respective parent's affinity. Dashed lines indicate LOESS regression (with 95\% confidence interval).}
\label{fig:locality}
\end{figure}

Figure \ref{fig:locality} illustrates the correlations found on the mean and standard deviation and are in line with expectations. It also appears that correlations are stronger in the deep network regime, which makes intuitive sense, as a single mutated value modifies a smaller proportion of the network.

\paragraph{Partial evaluation}
\label{par:assval_partialevaluation}
Here, we refer to partial evaluation as the training of candidate architectures with a subset of the training set and an aggressive early stopping policy, as presented in Section \ref{par:evaluation}. As in the previous experiment, we generate random architectures of depths 3 and 9 in the Fashion-MNIST search space and evaluate them under the partial evaluation regime. Next, we train them to convergence on the full dataset (training details in Section \ref{ssec:fullscaleexp}), from the inherited weights.

\begin{figure}[t]
\centering
\begin{minipage}{.48\textwidth}
  \centering
  \includegraphics[width=\linewidth]{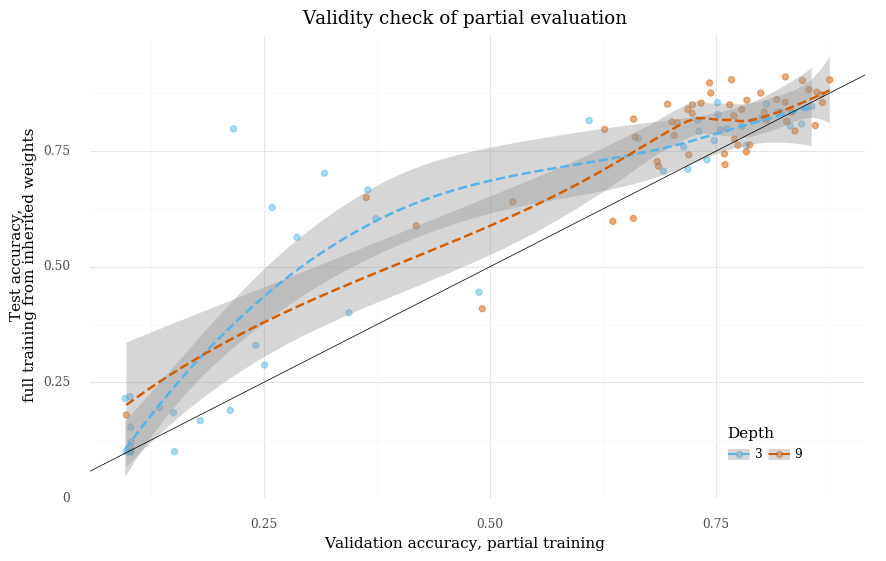}
  \captionof{figure}{Validity check of the partial evaluation strategy. Spearman rank correlation \textbf{overall}: $r=0.82, p<10^{-24}$;  \textbf{for depth 3}: $r=0.91, p < 10^{-19}$;  \textbf{for depth 9}: $r=0.65, p < 10^{-8}$. Dashed lines indicate LOESS regression (with 95\% confidence interval). Solid line  = Identity}
  \label{fig:partialevaluation}
\end{minipage}%
\hspace{0.5cm}
\begin{minipage}{.48\textwidth}
  \centering
  \includegraphics[width=\linewidth]{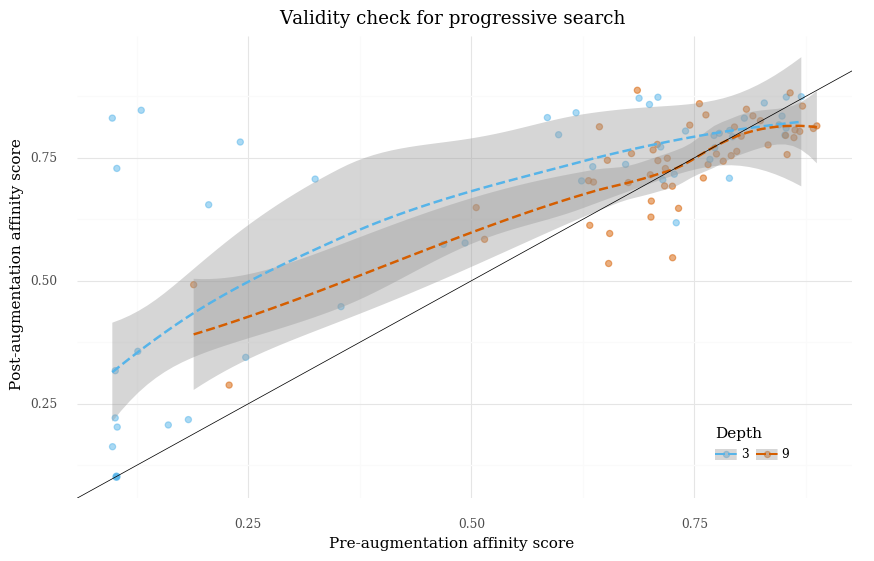}
  \captionof{figure}{Validity check of progressive search. Spearman rank correlation  \textbf{overall}: $r=0.64, p<10^{-13}$;  \textbf{for depth 3}: $r=0.66, p < 10^{-8}$;  \textbf{for depth 9}: $r=0.66, p < 10^{-8}$. Dashed lines indicate LOESS regression (with 95\% confidence interval). Solid line  = Identity}
  \label{fig:progressivesearch}
\end{minipage}
\end{figure}

We plot the full-training test accuracy values against the partial-training validation accuracy values in Figure \ref{fig:partialevaluation}. The correlation is very strong for shallow networks and somewhat weaker for deeper architectures. This can be explained by the fact that deeper architectures typically learn slower, therefore applying the same termination criteria to all network sizes might be suboptimal. Exploring alternative policies might constitute future work; nevertheless, the correlation is still robust even at depth 9, so that partial evaluation is still indicative of the final accuracy. Moreover, by the time a depth of 9 is reached, most networks perform reasonably well and they are going to be ensembled anyway so that it is less critical for partial evaluation to provide an exact indication of the ultimate performance -- which can be seen as a further benefit of our approach. Further experiments have also shown that these correlations are slightly weaker when weights are not inherited but reintialized, which justifies the combination of partial training and weight inheritance.

\paragraph{Progressive search}
\label{par:assval_progressivesearch}
Progressive search is the idea of incrementally building architectures by adding layers one by one. The underlying assumptions need to be verified by answering these two questions: 1) Does adding a layer to a strong candidate, in expectation, yield an even stronger candidate? 2) Given two candidates, does, in expectation, augmenting the stronger one yield a stronger candidate than augmenting the weaker one?

To test these assumptions, we again generate a population of random architectures of depths 3 and 9. We train them under the partial evaluation regime, measure their affinity score (validation accuracy), then augment each one them with an additional layer (with inherited weights in all other layers) and repeat the evaluation process. Figure \ref{fig:progressivesearch} addresses the two questions above: 1) Shallow networks almost always benefit from augmentation and while this is less true for deep networks, there is no catastrophic loss of performance due to augmentation; 2) there is significant correlation between affinity scores of networks before and after augmentation.

\subsection{Full-scale experiments}
\label{ssec:fullscaleexp}
Having verified the assumptions behind ImmuNeCS, we can now turn to full scale experiments on the Fashion-MNIST and the CIFAR-10 tasks. All experiments were run on NVidia 2080Ti GPUs.

\paragraph{Fashion-MNIST}
We use sequential architectures sampled from the Fashion-MNIST search space described in Section \ref{par:fashionmnistsearchspace} and Table \ref{tab:searchspace} (left). The meta-parameters used in the main experiment are:
\begin{itemize}
    \item Search: population size $N=12$ (the small population size is made possible by the relatively small search space and the progressive search strategy), number of clones per parent $n_c=3$, number of augmented copies per parent $n_a=3$, number of random networks inserted at each generation $N_i=2$, mutation factor $\rho=0.2$, patience $\pi=2$, threshold $\tau=0.0075$, maximum number of generations $= 20$;
    \item Partial evaluation: training set size $=12,000$ samples (20\% of the total training data available), validation set size $=12,000$ samples, batch size$=128$, data pre-processing: pad with zeros to 34x34 and crop back to 28x28, initial learning rate $=0.1$, optimizer: Adam with cosine annealing \cite{loshchilov2016} (no restart; the choice of cosine annealing is justified as in \cite{elsken2017}), $\beta_1=0.95$, $\beta_2=0.99$, weight decay $=10^{-5}$, early stopping patience $=2$ epochs, early stopping threshold $=0.005$, maximum number of epochs $=15$ (most evaluations stop around 10 epochs given these early stopping criteria);
    \item Final training: retain $N=12$ architectures; training hyperparameters are identical to partial evaluation except: training set size $=60,000$ samples, initial learning rate $=0.1/3$, 1 restart after 25 epochs, fixed number of epochs $=50$ (no early stopping).
\end{itemize}

The results obtained are summarized and compared to representative works are presented in Table \ref{tab:fmnistresults}. Our method achieves an average over 6 runs of 94.09\% (standard deviation 0.20\%), outperforming the others by a significant margin while requiring a comparable amount of resources. Admittedly, measuring resources in GPU-days is somewhat inaccurate because hardware improves over time; however this metric can still give us an estimate of each method's efficiency. 

\begin{table}
\centering
\begin{threeparttable}
\begin{tabular}{@{}lcccc@{}}
\toprule
\textbf{Model} & \textbf{Reported} & \textbf{Test accuracy \%} & \textbf{GPU-days} \\ \midrule
\begin{tabular}[c]{@{}l@{}} Auto-Keras \cite{jin2019} \end{tabular} & Best & 92.56 & 0.5 \\
\begin{tabular}[c]{@{}l@{}} NASH\tnote{a} \cite{elsken2017} \end{tabular} & Best & 91.95 & 0.5\\
\begin{tabular}[c]{@{}l@{}} Gradient Evolution\tnote{$\star$} \cite{mitschke2018} \end{tabular} & Best (Median) & 91.36 (90.58) &  NA  \\
\begin{tabular}[c]{@{}l@{}} DeepSwarm \cite{byla2019} \end{tabular} & Best (Mean) & 93.56 (93.25) & 1.2\\
\begin{tabular}[c]{@{}l@{}} \textbf{ImmuNeCS}\tnote{$\star$} \end{tabular} & \textbf{Best (Mean$\pm$std)} & \textbf{94.61 (94.37$\pm$0.15)} & \textbf{0.4} \\ 
\bottomrule
\end{tabular}
\begin{tablenotes}
    \footnotesize
    \item [a] As implemented in \cite{jin2019}.
    \item [$\star$] Evolutionary method.
\end{tablenotes}
\caption{Comparison of results of our method with others on Fashion-MNIST.}
\end{threeparttable}
\footnotetext{As implemented and reported in \cite{jin2019}}
\end{table}
\label{tab:fmnistresults}

\paragraph{CIFAR-10}
We sample architectures from the CIFAR-10 search space described in Section \ref{par:cifar10searchspace} and Table \ref{tab:searchspace} (left). The meta-parameters are the same as with Fashion-MNIST, with the following exceptions that are derived from common sense rather than lengthy tuning:
\begin{itemize}
    \item Given the larger number of possible operation types (and thus, the denser sampling space), we sample them from $\mathcal{N}(0, (\sigma/2)^2)$, keeping the definition of $\sigma$ provided in Section \ref{par:cloneandmutate}. All other items are sampled as before. For the same reason, we now generate 5 augmented copies per parent at augmentation, and somewhat offset this with reducing random insertions to 1. Given the expected slower progress of the search, we reduce the stopping threshold to 0.005.
    \item Partial evaluation: training set size $=10,000$ samples, validation set size $=10,000$ samples, batch size=$64$, data pre-processing: reflection padding to 40x40 and crop back to 32x32, cutout \cite{devries2017}, and random flip, initial learning rate $=0.05$, early stopping threshold $=0.003$, maximum number of epochs $=30$.
    \item Final training: to save on retraining time, we retain only the 5 best architectures ranked by affinity score; training hyperparameters are identical to partial evaluation except: training set size $=50,000$ samples, initial learning rate $=0.05/3$, 2 restarts after 30 and 90 epochs, fixed number of epochs $=210$ (no early stopping).
\end{itemize}

\begin{table}[t]
\centering
\begin{threeparttable}
\begin{tabular}{@{}lccc@{}}
\toprule
\textbf{Model} & \textbf{Reported} & \textbf{Test accuracy \%} & \textbf{GPU-days} \\ \midrule
\begin{tabular}[c]{@{}l@{}}ResNet-110\tnote{$\mathparagraph$} $\;$  \cite{he2016} \end{tabular} & Best (Mean$\pm$std) & 93.57 (93.39$\pm$0.16) & --  \\ \midrule
\begin{tabular}[c]{@{}l@{}}NAS depth 15 \cite{zoph2017}\tnote{$\dagger$} \end{tabular} & Best & 94.50 & 12600  \\
\begin{tabular}[c]{@{}l@{}}NAS depth 39, extra filters\tnote{$\dagger$}\end{tabular} & Best & 96.35 & 12600 \\  \midrule
\begin{tabular}[c]{@{}l@{}}NASNet-A 28M \cite{zoph2018}\tnote{$\ddagger$$\diamond$} \end{tabular} & Best & 97.60 & 1800  \\ 
\begin{tabular}[c]{@{}l@{}}NASNet-A 3.3M\tnote{$\ddagger$$\diamond$}\end{tabular} & Best & 97.35 & 1800 \\ \midrule
\begin{tabular}[c]{@{}l@{}}Large-Scale Evo. \cite{real2017}\tnote{$\star$} \end{tabular} & Best (Mean$\pm$std) & 94.60 (94.10$\pm$0.40) & 3000 \\
\begin{tabular}[c]{@{}l@{}}Large-Scale Evo. ensemble\tnote{$\star$} \end{tabular} & Best & 95.60 & 3000 \\ \midrule
\begin{tabular}[c]{@{}l@{}}AmoebaNet-B 2.8M \cite{real2019}\tnote{$\ddagger$$\diamond$$\star$}\end{tabular} & Mean$\pm$std & 97.45$\pm$0.05 & 4500 (TPU)  \\
\begin{tabular}[c]{@{}l@{}}AmoebaNet-B 34M\tnote{$\ddagger$$\diamond$$\star$}\end{tabular} & Mean$\pm$std & 97.87$\pm$0.04 & 4500 (TPU) \\  \midrule
\begin{tabular}[c]{@{}l@{}}NASH single model \cite{elsken2017} \end{tabular} & Mean & 94.80 & 1 \\
\begin{tabular}[c]{@{}l@{}}NASH snapshot ensemble\end{tabular} & Mean & 95.30 & 2 \\ \midrule
\begin{tabular}[c]{@{}l@{}}LEMONADE SS-I \cite{elsken2019b}\tnote{$\star$} \end{tabular} & Best & 96.50 & 56 \\
\begin{tabular}[c]{@{}l@{}}LEMONADE SS-II\tnote{$\ddagger$$\diamond$$\star$}\end{tabular} & Best & 96.60 & 56 \\ \midrule
\begin{tabular}[c]{@{}l@{}}ENAS Macro \cite{pham2018}\end{tabular} & Best & 95.77 & 0.3 \\
\begin{tabular}[c]{@{}l@{}}ENAS Micro\tnote{$\ddagger$$\diamond$}\\ \end{tabular} & Best & 97.11 & 0.5 \\ \midrule
\begin{tabular}[c]{@{}l@{}}DARTS 1st Order\cite{liu2018}\tnote{$\ddagger$$\diamond$}\end{tabular} & Best & 97.05 & 1.5 \\
\begin{tabular}[c]{@{}l@{}}DARTS 2nd Order\tnote{$\ddagger$$\diamond$}\end{tabular} & Mean$\pm$std & 97.17$\pm$0.06 & 4 \\ \midrule
\begin{tabular}[c]{@{}l@{}}CGP-CNN \cite{suganuma2017}\tnote{$\star$} \end{tabular} & Best (Mean) & 94.34 (93.95) & 27 \\ \midrule
\begin{tabular}[c]{@{}l@{}}PNAS \cite{liuc2017}\tnote{$\dagger$$\diamond$}\end{tabular} & Mean$\pm$std & 96.59$\pm$0.09 & 300 \\ \midrule
\textbf{ImmuNeCS}\tnote{$\star$} & \textbf{Best (Mean$\pm$std)} & \textbf{95.62 (94.97$\pm$0.50)} & \textbf{14} \\ 
\bottomrule
\end{tabular}
\begin{tablenotes}
    \footnotesize
    \item [$\mathparagraph$] Hand-crafted. $^{\dagger}$ Hyperparameter search before final training. $^{\ddagger}$ Model augmentation before final training. $\diamond$ Repeated cell-based method. $^{\star}$ Evolutionary method.
\end{tablenotes}
\caption{Comparison of results of our method with others on the CIFAR-10 task.}
\end{threeparttable}
\end{table}
\label{tab:cifar10results}

The initial results and comparisons on this task are summarized in Table \ref{tab:cifar10results}. Although further experiments are required, ImmuNeCS seems to achieve a competitive balance of performance and efficiency, particularly among methods that do not use a cell-based search space. We point out that the search space has not been optimized in any way, we simply reproduced six blocks from traditional hand-crafted architectures. It is very likely that a more thorough analysis of appropriate block candidates would improve these results significantly. One could even imagine a cell-based approach to design a few blocks, followed by the same high-level architecture search as above to combine these blocks (similar to \cite{miikkulainen2017}). Moreover, where a number of papers apply some form of post-processing specifically at the final retraining stage (hyperparameter search, augmentation etc.), in our method, the final architectures found by the algorithm are retrained without any modification nor hyperparameter optimization.

\subsection{Comparison with Random Search}

Some NAS methods have come under question because their superiority over Random Search (RS) could not be established in rigourous experiments \cite{sciuto2019, li2019}. This might indicate that the techniques they use add complexity without helping performance, and that the performance mostly comes from the designed search spaces rather than the search algorithms. It is therefore important to compare ImmuNeCS with RS to ensure this is not the case with our approach. To this end, we run ImmuNeCS seven times on Fashion-MNIST at different evaluation budgets, which are obtained by varying the search-stopping threshold $\tau$ and the maximum number of generations. We then run a full RS at different budget values spanning the range covered by ImmuNeCS.

The number of layers of architectures is important to their performance, therefore we should pay attention to how it is sampled in RS. The minimum value is easy enough to decide; we set it at the starting number of layers in our Fashion-MNIST experiments (3). Setting a maximum value requires more thought because there is no maximum depth defined in our method other than what is allowed by the maximum number of generations, 40 in this case. We could use the highest value used by ImmuNeCS, but this artificially biases the search into a domain that has been previously discovered by the AIS rather than hard-coded. This would be unfair to our algorithm as it lets RS benefits from one of its key contributions, namely its ability to find minimal architectures. On the other hand, using a very high maximum number of layers enormously expands the search space, making it harder for RS to perform well within a given budget, which would also be unfair. Considering that the best architectures found by ImmuNeCS have 8 layers, we therefore choose to let random search sample uniformly from the range $[3,12]$. In practice, RS's best performing architectures have 10 layers, which highlights the superior efficiency of our method.

Random search achieves $93.37 \pm 0.35\%$ in test accuracy for the ensemble, significantly worse than our method (one-tailed $p<0.001$).

\subsection{Comparison to Genetic Algorithm} 
In \cite{real2017}, Real \textit{et al.} implement an evolutionary algorithm that is akin to GA \cite{holland1992} without crossover. Unlike CLONALG, it samples its mutations from a uniform distribution. Candidates are selected by pairwise tournament selection.

We reproduce this search algorithm and adapt it to our approach. In particular, we use the same population size, partial evaluation regime, early stopping mechanism, final retraining, and ensembling procedure as in ImmuNeCS. We run it under the same evaluation budget (smaller-scale setting). The final committee's mean test accuracy is 93.95\%$\pm$0.4 with a best result of 94.41\%, significantly below ImmuNeCS (one-tailed $p=0.039$). While the accuracy of the best individual model is very similar on average (93.39\% for ImmuNeCS vs. 93.31\% for GA), the gain from ensembling is higher in ImmuNeCS (0.8\%pt vs. 0.6\%pt on average). These results indicate that using the AIS brings benefits, presumably in terms of diversity, that help ensembling.

\subsection{Transferability}
An important benefit of the cell-based approach to NAS is that it demonstrates good transferability to different tasks. Indeed, it is easy to assemble a larger network than the one used during search, simply by stacking more cells. It is less obvious how to achieve such transferability in full-architecture search methods. We hypothesize that the plurality of ImmuNeCS's solutions helps them in that respect, allowing them to achieve competitive results even when the architecture search was conducted on an easier task. We run two experiments to verify this.

\paragraph{MNIST to Fashion-MNIST}
We first run ImmuNeCS on MNIST, a very simple task by modern standards, using the same search space and search parameters as described in Sections \ref{par:fashionmnistsearchspace} and \ref{ssec:fullscaleexp} for the Fashion-MNIST task. We obtain an NNC of 12 CNNs that are typically shallower than those obtained through a direct search on the Fashion-MNIST task. We train them first on MNIST and obtain a final accuracy of 99.6\%. We then train the same architectures on Fashion-MNIST and measure a final accuracy of 92.8\%, which is still among the best results presented in Table \ref{tab:fmnistresults}. Interestingly, the performance ranking of architectures on MNIST is completely different to that on Fashion-MNIST, indicating that a search focused on the single best model would be unlikely to return the best architecture for both tasks.

\paragraph{CIFAR-10 to CINIC-10}
In a second experiment, we take the population found in our CIFAR-10 experiment and train it on CINIC-10 \cite{darlow2018}, combining the training and validation sets, i.e. 180,000 samples. CINIC-10 is an image dataset designed to be a direct drop-in replacement for CIFAR-10, but much more challenging as it mixes images from CIFAR-10 with downsized images from ImageNet, meaning that it contains images from different distributions to that on which the search was performed. Moreover, its images are noisier and can mix elements from different classes within the same image. After retraining using the same procedure as described in Section \ref{ssec:fullscaleexp} for CIFAR-10, we obtain a committee accuracy of 88.72\% on the test set (90,000 samples), with a 1.36\%pt gain from ensembling. This places the performance between VGG-16 \cite{simonyan2014} (87.77\%) and ResNet-18 \cite{he2016} (90.27\%), two commonly used hand-crafted models. 

\section{Conclusion and Discussion}
We have presented ImmuNeCS, a novel approach to the automatic design of deep learning systems. Instead of focusing on producing a single architecture, we promote a diverse population of competent models that are ensembled to achieve competitive results with reasonable resource requirements. To improve efficiency, we use techniques whose underlying assumptions are verified through dedicated experiments. We also show that our method outperforms random search in a fair comparison, and remark that the NNC found on a given task is able to generalize to a more complex task.

Furthermore, our method presents other benefits: 1) It is conceptually simple and therefore approachable from non-experts, 2) it does not enforce an architecture based on repeated identical cells, and is thus able to discover irregular patterns, 3) it is flexible and can accommodate many search spaces. One obvious drawback is that inference is slower, as we need to aggregate the predictions of all members of the NNC rather than a single model. However, in applications where predictions are not required in real-time, this could be an acceptable trade-off. Another limitation is that the final retraining phase involves training several neural networks rather than a single one. Again, we believe that this one-off cost is acceptable in many use-cases.

Future work includes expanding the CIFAR-10 search space to further improve performance, and running experiments on tasks from the field of medical imaging to assess ImmuNeCS's flexibility and usefulness in real-life situations. In order to make the algorithm more efficient, we also intend to investigate methods that could help guide the search towards promising areas of the search space.

\bibliographystyle{unsrt}  
\bibliography{references.bib}  

\end{document}